\documentclass[10pt,twocolumn,final]{article}
\pdfoutput=1

\usepackage[top=2cm, left=1.5cm, bottom=2cm, right=1.5cm]{geometry}

\usepackage{amsmath}
\usepackage{amssymb}
\usepackage{latexsym}

\usepackage{graphicx}
\usepackage{balance}

\usepackage[square,numbers,sort&compress]{natbib}

\usepackage{fancyhdr}
\makeatother
\usepackage{url}
\usepackage{xcolor}
\usepackage{hyperref}
\usepackage{pdfpages}

\usepackage{amssymb}
\usepackage{amsmath}
\usepackage{bm}

\newcommand{\fscore}[1]{F\textsubscript{#1}-score}
\newcommand{\fscores}[1]{F\textsubscript{#1}-scores}

\usepackage{caption}
\newcommand{\customCaption}[2]{
\vspace{-15pt}
\begin{flushleft}
\footnotesize
\textbf{Figure~#1}: #2
\end{flushleft}
}

\newcommand{\customCaptionTable}[2]{
\vspace{-15pt}
\begin{flushleft}
\footnotesize
\textbf{Table~#1}: #2
\end{flushleft}
}

\begin{document}

\twocolumn[
  \begin{@twocolumnfalse}

\title{
Rotation Invariance and Extensive Data Augmentation:\newline a strategy for the MItosis DOmain Generalization (MIDOG) Challenge
}
\date{}

\vspace*{-50pt}
\begin{minipage}{\textwidth}
\centering
\author{
Maxime W. Lafarge and Viktor H. Koelzer
}
\end{minipage}

\maketitle

\vspace*{-20pt}
\begin{center}
\begin{minipage}{0.9\textwidth}
\begin{flushleft}
\footnotesize
\textit{Department of Pathology and Molecular Pathology, University Hospital and University of Zurich, Zurich, Switzerland} 
\end{flushleft}
\end{minipage}
\end{center}

\vspace*{20pt}
\begin{abstract}
Automated detection of mitotic figures in histopathology images is a challenging task: here, we present the different steps that describe the strategy we applied to participate in the MIDOG 2021 competition.
The purpose of the competition was to evaluate the generalization of solutions to images acquired with unseen target scanners (hidden for the participants) under the constraint of using training data from a limited set of four independent source scanners.
Given this goal and constraints, we joined the challenge by proposing a straight-forward solution based on a combination of state-of-the-art deep learning methods with the aim of yielding robustness to possible scanner-related distributional shifts at inference time.
Our solution combines methods that were previously shown to be efficient for mitosis detection: hard negative mining, extensive data augmentation, rotation-invariant convolutional networks.

We trained five models with different splits of the provided dataset. The subsequent classifiers produced \fscore{1} with a mean and standard deviation of $0.747 {\pm} 0.032$ on the test splits. The resulting ensemble constitutes our candidate algorithm: its automated evaluation on the preliminary test set of the challenge returned a \fscore{1} of $0.6828$.
\end{abstract}
\vspace*{40pt}

  \end{@twocolumnfalse}
]

\subsection*{Dataset Preparation}
\label{preparation}
\noindent
The organizers of MIDOG 2021 \citep{midog2021} provided annotated images from $150$ cases ($50$ cases each from $3$ different source scanners).
$50$ images from a fourth scanner were provided but we chose not to use them in order to present a solution based solely on a supervised learning framework, thus leaving room for improvements for future work.

We created five folds of three splits such that we were able to train and validate multiple models with varying data distributions.
For each fold, we partitioned cases in splits with the following distribution: training(80\%), validation(10\%) and test(10\%), such that the distribution of scanners was identical within each split.
With this partition we intended to use as much available source data as possible for training while keeping a small proportion for internal validation and model selection.

\vspace{10pt}
\begin{figure}[hb!]
\begin{center}
\includegraphics[width=\columnwidth, trim=5pt 35pt 335pt 5pt, clip]{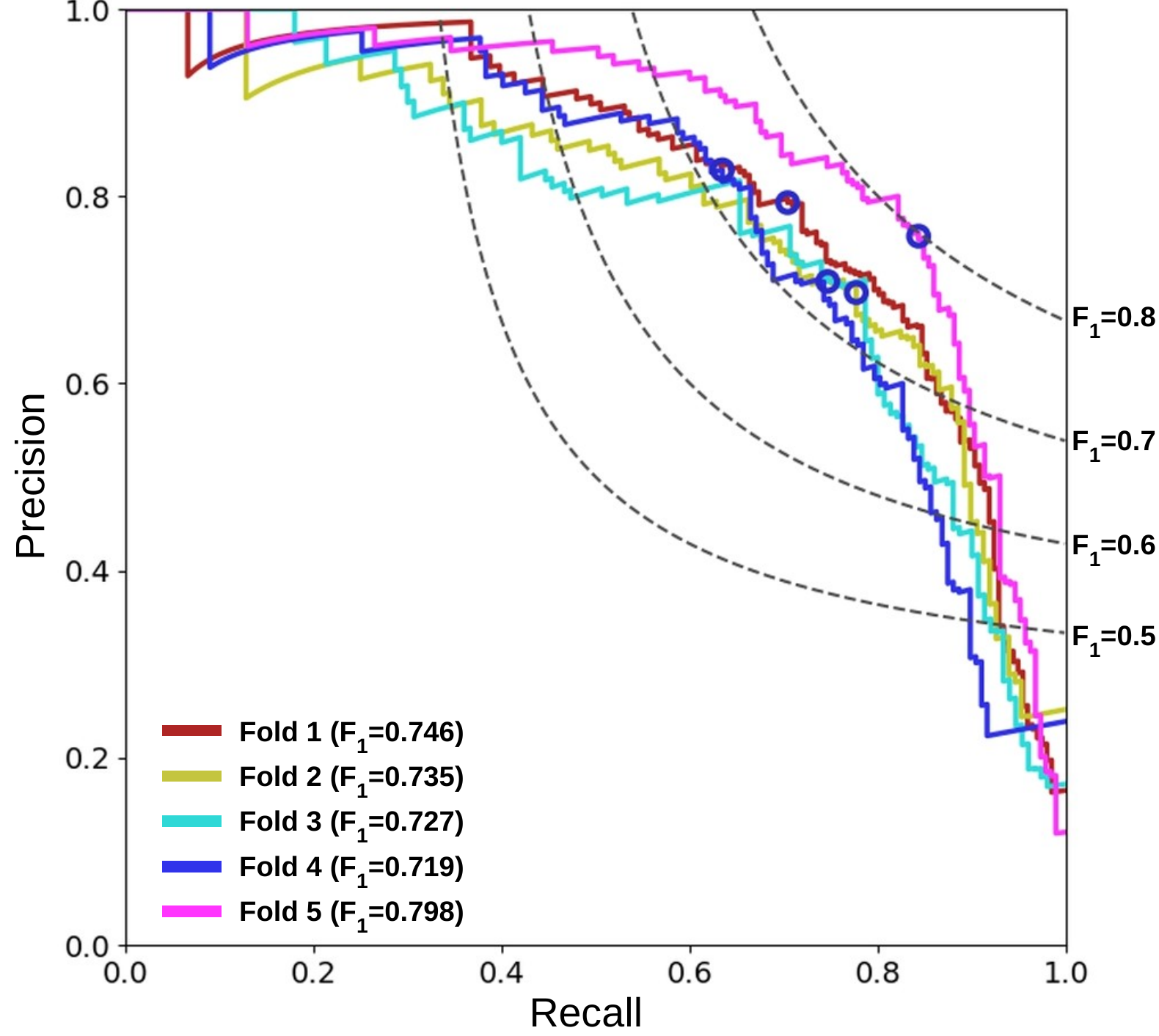}
\caption{} 
\label{fig:PRcurve}
\customCaption{\ref{fig:PRcurve}}{
Precision-Recall analysis of five models trained and evaluated on the different test sets for each fold of the dataset.
Dark blue circles show the performances achieved by the models using the operating points that maximized the \fscore{1} on the validation sets.
}
\end{center}
\end{figure}

\subsection*{Model Architecture}
\label{architecture}
We modeled the conditional likelihood of the mitosis class given an input image patch of size $77{\times}77$ at magnification $40{\times}$ using convolutional neural networks (CNNs).
Motivated by the benefits of rotation invariance properties of deep learning models for computational pathology tasks \citep{bekkers2018roto,veeling2018rotation,lafarge2020roto,graham2020dense}, we used roto-translation equivariant convolutional layers with a $8$-fold discretization of the orientation axis \citep{lafarge2020roto}.
As this structure guarantees the roto-translation equivariance of the internal activations and invariance of the output of the models with respect to the orientation of the input, rotation augmentation at training and inference time becomes an unnecessary step.
Furthermore, this gained invariance property prevents learning possible biases related to the orientation of the images.
\newline
The detailed architecture we used is described in Table \ref{tab:modelArchitecture}.

\subsection*{Training Procedure \newline and Data Augmentation}
\label{training}

We trained our models with batches of size $64$ balanced between mitotic figures and non-mitotic objects, and optimized the weights of the models via minimization of the cross-entropy loss.
We used the \textit{Adam} optimizer (learning rate $3{\times}10^{-4}$), with a step-wise decay by a factor $0.8$ every $5000$ iterations, and stopped training after convergence of the training loss. We used weight decay with coefficient $2{\times}10^{-4}$. For inference time, we kept the weights of the model that achieved the minimum validation loss.

\begin{figure}[h!]
\begin{center}
\includegraphics[width=0.95\columnwidth, trim=5pt 160pt 50pt 5pt, clip]{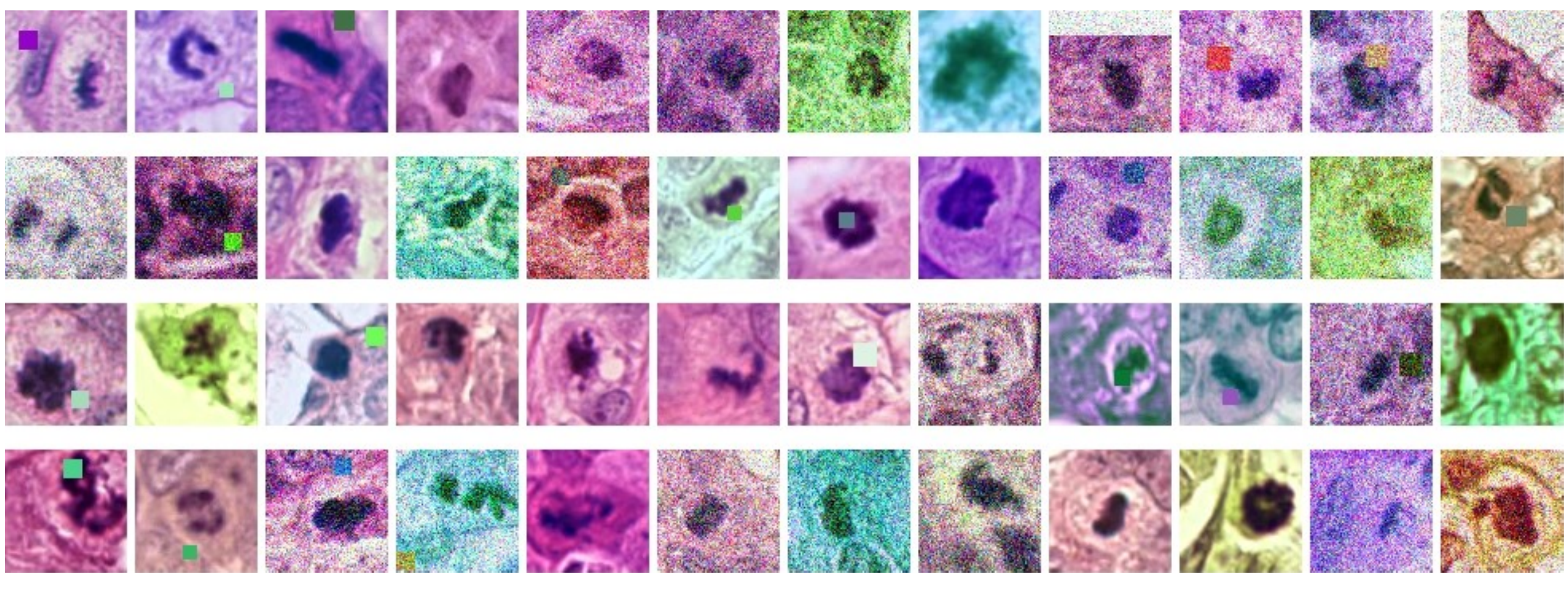}
\end{center}
\vspace{-10pt}
\caption{} 
\label{fig:dataAugmentation}
\customCaption{\ref{fig:dataAugmentation}}{
Example of mitosis-centered image patches transformed according to our random data augmentation protocol.
}
\end{figure}

In order to ensure the generalization of our model to variations of appearance related to unseen scanners, we opted for an extensive and aggressive data augmentation strategy.
For this purpose, we applied a series of random transformations according to the protocol described in Table \ref{tab:dataAugmentation}.
Examples of transformed image patches are shown in Figure \ref{fig:dataAugmentation}.
This approach is motivated by related works showing the effectiveness of data augmentation for mitosis detection \citep{tellez2018heAugmentation,lafarge2019domain}.

\begin{table}
\caption{} 
\label{tab:modelArchitecture}
\customCaptionTable{\ref{tab:modelArchitecture}}{
Architecture of the CNN used in this work.
Shape of output tensors are written with the following format: $\text{(\textit{Orientations}}{\times}\text{)}\text{\textit{Channels}}\text{(}{\times}\text{\textit{Height}}{\times}\text{\textit{Width})}$.
\newline Shape of operator tensors are written with the following format: $\text{(\textit{Orientations}}{\times}\text{)}\text{\textit{Out.Ch.}}{\times}\text{\textit{In.Ch.}}{\times}\text{\textit{Ker.Height}}{\times}\text{\textit{Ker.Width}}$. 
\newline {*} indicates that the operation is followed by a \textit{Batch Normalization} layer and a leaky \textit{ReLU} non-linerarity (coefficient $0.01$).
}
\begin{center}

\renewcommand{\arraystretch}{0.9}
\footnotesize
\begin{tabular}{p{0.4\columnwidth} p{0.25\columnwidth} p{0.2\columnwidth}}
\textit{Layer}
& \textit{Operator Shape}
& \textit{Output Shape} \\\hline\hline
Input               & --                                        & $3{\times}77{\times}77$ \\\hline
Lifting Convolution {*}  & $16{\times}3{\times}4{\times}4$           & $8{\times}16{\times}74{\times}74$ \\\hline
Max Pooling         & $2{\times}2$                              & $8{\times}16{\times}37{\times}37$ \\\hline
SE(2,8)-Convolution {*} & $8{\times}16{\times}16{\times}4{\times}4$ & $8{\times}16{\times}34{\times}34$ \\\hline
Max Pooling         & $2{\times}2$                              & $8{\times}16{\times}17{\times}17$ \\\hline
SE(2,8)-Convolution {*} & $8{\times}16{\times}16{\times}4{\times}4$ & $8{\times}16{\times}14{\times}14$ \\\hline
Max Pooling         & $2{\times}2$                              & $8{\times}16{\times}7{\times}7$ \\\hline
SE(2,8)-Convolution {*} & $8{\times}16{\times}16{\times}4{\times}4$ & $8{\times}16{\times}4{\times}4$ \\\hline
SE(2,8)-Convolution {*} & $8{\times}32{\times}16{\times}4{\times}4$ & $8{\times}32{\times}1{\times}1$ \\\hline
Maximum Projection  & --                                        & $32$ \\\hline
Fully Connected {*}     & $64{\times}32$                        & $64$ \\\hline
Fully Connected + Sigmoid & $1{\times}64$                       & $1$ \\\hline

\end{tabular}
\end{center}
\end{table}

\begin{table}[h!]
\caption{} 
\label{tab:dataAugmentation}
\customCaptionTable{\ref{tab:dataAugmentation}}{
Data augmentation protocol: for each input image patch, we scanned the following list of transformations and applied it with a given probability, after random sampling of a set of coefficients.
}

\begin{center}
\renewcommand{\arraystretch}{0.9}
\footnotesize
\begin{tabular}{p{0.45\columnwidth} p{0.3\columnwidth} p{0.12\columnwidth}}
\textit{Transformation}
& \textit{Coefficients}
& \textit{Probability} \\\hline\hline

Transposition & -- & $50\%$ \\\hline
Color Shift & $c_{r,g,b} \sim U\left[-13, 13\right]$ & $50\%$ \\\hline
Gamma Correction & $\gamma_{r,g,b} \sim U\left[0.9, 1.5\right]$ & $50\%$ \\\hline
Hue Rotation & $h \sim U\left[0, 1\right]$ & $50\%$ \\\hline
Spatial Shift & $\Delta_{x,y} {\sim} U\left[-12\text{px}, 12\text{px}\right]$ & $100\%$ \\\hline
Spatial Scale & $\alpha \sim  U\left[-13\%, 13\%\right]$ & $50\%$ \\\hline
Additive Gaussian Noise & $c_{x,y} \sim \mathcal{N}(0, 50)$ & $50\%$ \\\hline
Cutout\citep{devries2017cutout} (random color/size $s$) & $s \sim U\left[8\text{px}, 16\text{px}\right]$ & $50\%$ \\\hline

\end{tabular}
\end{center}
\end{table}

Generating training batches via random sampling of non-mitotic image patches is known to be a suboptimal approach for mitosis detection as models are less exposed to challenging non-mitotic objects during training \citep{cirecsan2013mitosis}.
Therefore, to encourage the model to discriminate challenging non-mitotic objects, for each fold, we sequentially resampled the dataset by removing easy classified patches using a protocol derived from \citet{cirecsan2013mitosis} using first versions of the models trained via random sampling of the training sets.

\subsection*{Inference Time}
\label{inference}
At inference time, the fully convolutional structure of our models enables their dense application on large test images which produces probability maps.
Candidate mitotic figures are identified as local maxima after applying non-maxima suppression within a radius of $30$px.
Our models are then turned into binary classifiers by setting a cutoff threshold (operating point) that is selected such that the \fscores{1} on the validation sets were maximized.
We applied this procedure to generate a classifier for each fold, and then gathered the $5$ models to form an ensemble.
The performance of these classifiers on the source test sets are reported in Figure \ref{fig:PRcurve}.
For new test images the detections of each classifier are considered as votes for candidate mitoses and we filter out detections that get less than $2$ votes.

\subsection*{Conclusions and Discussion}
\label{conclusion}
We proposed a straight-forward approach combining multiple state-of-the-art solutions to tackle the generalization problem for scanner-related distributional shifts in the context of the MIDOG2021 competition.
We report a lower performance of our solution on the preliminary test set provided by the organizers compared to the performances we obtained on the source test sets, suggesting that the generalization of our model is limited to some extent.
We hope that our methodology can be considered as a baseline, that could potentially be improved using additional training components for domain generalization.
In future work, we will aim at investigating the reasons of the generalization limitations of the presented method.

\subsection*{References}
\vspace*{20pt}
\bibliographystyle{unsrtnat}
\renewcommand\refname{}
\vspace*{-40pt}
\footnotesize
\bibliography{references}

\begin{thebibliography}{9}
\providecommand{\natexlab}[1]{#1}
\providecommand{\url}[1]{\texttt{#1}}
\expandafter\ifx\csname urlstyle\endcsname\relax
  \providecommand{\doi}[1]{doi: #1}\else
  \providecommand{\doi}{doi: \begingroup \urlstyle{rm}\Url}\fi

\bibitem[Aubreville et~al.(2021)Aubreville, Bertram, Veta, Klopfleisch,
  Stathonikos, Breininger, ter Hoeve, Ciompi, and Maier]{midog2021}
Marc Aubreville, Christof Bertram, Mitko Veta, Robert Klopfleisch, Nikolas
  Stathonikos, Katharina Breininger, Natalie ter Hoeve, Francesco Ciompi, and
  Andreas Maier.
\newblock Mitosis domain generalization challenge.
\newblock \emph{Zenodo, doi: 10.5281/zenodo.4573978}, 2021.

\bibitem[Bekkers et~al.(2018)Bekkers, Lafarge, Veta, Eppenhof, Pluim, and
  Duits]{bekkers2018roto}
Erik~J Bekkers, Maxime~W Lafarge, Mitko Veta, Koen~AJ Eppenhof, Josien~PW
  Pluim, and Remco Duits.
\newblock Roto-translation covariant convolutional networks for medical image
  analysis.
\newblock In \emph{Proceedings of the International Conference on Medical Image
  Computing and Computer-Assisted Intervention (MICCAI)}, volume 11070, pages
  440--448, 2018.

\bibitem[Veeling et~al.(2018)Veeling, Linmans, Winkens, Cohen, and
  Welling]{veeling2018rotation}
Bastiaan~S Veeling, Jasper Linmans, Jim Winkens, Taco Cohen, and Max Welling.
\newblock Rotation equivariant cnns for digital pathology.
\newblock In \emph{Proceedings of the International Conference on Medical Image
  Computing and Computer-Assisted Intervention (MICCAI)}, pages 210--218, 2018.

\bibitem[Lafarge et~al.(2021)Lafarge, Bekkers, Pluim, Duits, and
  Veta]{lafarge2020roto}
Maxime~W Lafarge, Erik~J Bekkers, Josien~PW Pluim, Remco Duits, and Mitko Veta.
\newblock Roto-translation equivariant convolutional networks: Application to
  histopathology image analysis.
\newblock \emph{Medical Image Analysis}, 68:\penalty0 101849, 2021.

\bibitem[Graham et~al.(2020)Graham, Epstein, and Rajpoot]{graham2020dense}
Simon Graham, David Epstein, and Nasir Rajpoot.
\newblock Dense steerable filter cnns for exploiting rotational symmetry in
  histology images.
\newblock \emph{IEEE Transactions on Medical Imaging}, 39:\penalty0 4124--4136,
  2020.

\bibitem[Tellez et~al.(2018)Tellez, Balkenhol, Karssemeijer, Litjens, van~der
  Laak, and Ciompi]{tellez2018heAugmentation}
David Tellez, Maschenka Balkenhol, Nico Karssemeijer, Geert Litjens, Jeroen
  van~der Laak, and Francesco Ciompi.
\newblock H and e stain augmentation improves generalization of convolutional
  networks for histopathological mitosis detection.
\newblock In \emph{Proceedings of SPIE Medical Imaging}, page 105810Z, 2018.

\bibitem[Lafarge et~al.(2019)Lafarge, Pluim, Eppenhof, and
  Veta]{lafarge2019domain}
Maxime Lafarge, Josien Pluim, Koen Eppenhof, and Mitko Veta.
\newblock Learning domain-invariant representations of histological images.
\newblock \emph{Frontiers in Medicine}, 6:\penalty0 162, 2019.

\bibitem[DeVries and Taylor(2017)]{devries2017cutout}
Terrance DeVries and Graham~W Taylor.
\newblock Improved regularization of convolutional neural networks with cutout.
\newblock \emph{arXiv preprint arXiv:1708.04552}, 2017.

\bibitem[Cire{\c{s}}an et~al.(2013)Cire{\c{s}}an, Giusti, Gambardella, and
  Schmidhuber]{cirecsan2013mitosis}
Dan~C Cire{\c{s}}an, Alessandro Giusti, Luca~M Gambardella, and J{\"u}rgen
  Schmidhuber.
\newblock Mitosis detection in breast cancer histology images with deep neural
  networks.
\newblock In \emph{Proceedings of the International Conference on Medical Image
  Computing and Computer-Assisted Intervention (MICCAI)}, pages 411--418, 2013.

\end{thebibliography}
\balance


\end{document}